\documentclass[
twocolumn,
]{ceurart}

\usepackage[utf8]{inputenc}

\hypersetup{breaklinks=true}
\begin{document}

\copyrightyear{2022}
\copyrightclause{Copyright for this paper by its authors.
  Use permitted under Creative Commons License Attribution 4.0
  International (CC BY 4.0).}

\conference{MRC 2022 – The Thirteenth International Workshop Modelling | Reasoning | Context. Held at IJCAI-ECAI 2022, July 23-29, 2022, Vienna, Austria.}

\title{Learning to Represent Individual Differences for Choice Decision Making}
\title[mode=sub]{Modified for use with the MRC workshop series}


\author{Yan-Ying Chen}[%
  email=yan-ying.chen@tri.global,
]

\author{Yue Weng}[%
  email=yue.weng@tri.global,
]

\author{Alexandre Filipowicz}[%
  email=alex.filipowicz@tri.global,
]

\author{Rumen Iliev}[%
  email=rumen.iliev@tri.global,
]

\author{Francine Chen}[%
  email=francine.chen@tri.global,
]

\author{Shabnam Hakimi}[%
  email=shabnam.hakimi@tri.global,
]

\author{Yanxia Zhang}[%
  email=yanxia.zhang@tri.global,
]

\author{Matthew Lee}[%
  email=matthew.lee@tri.global,
]

\author{Kent Lyons}[%
  email=kent.lyons@tri.global,
]

\author{Charlene Wu}[%
  email=charlene.wu@tri.global,
]

\address{Toyota Research Institute, 4440 El Camino Real, Los Altos, CA 94022, United States}

\begin{abstract}
Human decision making can be challenging to predict because decisions are affected by a number of complex factors. Adding to this complexity, decision-making processes can differ considerably between individuals, and methods aimed at predicting human decisions need to take individual differences into account. Behavioral science offers methods by which to measure individual differences (e.g., questionnaires, behavioral models), but these are often narrowed down to low dimensions and not tailored to specific prediction tasks. This paper investigates the use of representation learning to measure individual differences from behavioral experiment data. Representation learning offers a flexible approach to create individual embeddings from data that are both structured (e.g., demographic information) and unstructured (e.g., free text), where the flexibility provides more options for individual difference measures for personalization, e.g., free text responses may allow for open-ended questions that are less privacy-sensitive. In the current paper we use representation learning to characterize individual differences in human performance on an economic decision-making task. We demonstrate that models using representation learning to capture individual differences consistently improve decision predictions over models without representation learning, and even outperform well-known theory-based behavioral models used in these environments. Our results propose that representation learning offers a useful and flexible tool to capture individual differences.
\end{abstract}


\maketitle

\section{Introduction}

There is a growing framework of machine learning models aimed at predicting human decisions \cite{plonsky2017,hartford2016}. However, predicting human decision making can be challenging because of complex factors that influence human choices. Indeed, factors such as uncertainty, risk, time, and perceived effort have all been shown to influence everyday decisions \cite{kahneman2013prospect,odum2011delay,kivetz2003effects}. In addition to these external factors, stable characteristics that distinguish one individual from another, known in psychology as "individual differences", can also interact with external factors to influence choices. For example, individual differences such as age, country of origin, education and factors related to mental health have all been shown to influence the extent to which people perceive and respond to risk when making decisions \cite{weber1998cross,mata2011age,gachter2021individual,hartley2012anxiety}. Thus, in order to accurately predict people's choices, modeling methods should be well suited to account for individual differences.

Despite the importance of individual differences in predicting human decisions, accurately measuring these differences can come with some challenges. Some individual differences, such as geographic location or age, are reasonably straight forward to measure. Others related more to preferences or personality traits (e.g., extroversion, pro-social tendencies, etc) can be more challenging. Current behavioral science approaches often use psychometric measures (e.g., questionnaires) to measure these latter types of individual differences. Properly validated questionnaires can be a powerful measure of different facets of personality. However, these are generally restricted to a low dimension and are not optimized toward the designated task, e.g., predicting choice decisions.




The current paper attempts to measure individual differences in human decisions using representation learning. Representation learning has been used to personalize predictions based on user data (e.g., personalize the results of a recommendation systems \cite{cheng2016}), and thus offers a rich class of methods to capture individual differences in the context of behavioral experiments.

We see two main advantages of representation learning over more traditional approaches to measuring individual differences. First, representation learning can be trained on a wide range of data types ranging from choices, to demographics, to open-ended text responses. This flexibility offers it as a method to capture individual differences from a broad range of data sources including some with higher variance (e.g., free text) and to fit use cases that need different levels or aspects of privacy for personalization. For example, in a use case where demographic information is sensitive, it can design relevant open-ended questions that allow participants to respond more comfortably using free text. Second, representation learning represents individuals in a high dimensional space that captures more complexity than more targeted and lower dimensional measures (e.g., questionnaires). This offers the ability to capture a richer representation of differences between individuals that are not limited to one modality. This, coupled with the broad data types to which representation learning is suited, offer it as a strong candidate method to measure individual differences.

This paper will investigate ways to leverage the strengths of representation learning to represent individual differences in human decision making. We compare our approach with a theory-based behavioral model \cite{kahneman2013prospect} specifically designed for modeling risky behaviors in gambling decision tasks such as the data used in our experiment. Our article makes three main contributions: 
\begin{enumerate}
\item we introduce the use of representation learning to model individual differences in behavioral experiments of decision making.
\item we lay out a novel framework to measure individual differences in decision-making from unstructured and open-ended text responses.
\item we investigate whether machine learning approaches can obtain competitive results in comparison to a model driven by well-known behavioral theories for the designated behavioral experiment. 
\end{enumerate}
In the following sections, we first discuss the related work followed by the dataset and problem definition. We then introduce the approaches used in representation learning, and our novel use of open-ended questions as well as the model driven by behavioral theories for comparison. We conclude with by presenting our experimental analysis and findings.

\begin{figure*}[t]
    \centering
    \includegraphics[width=14cm]{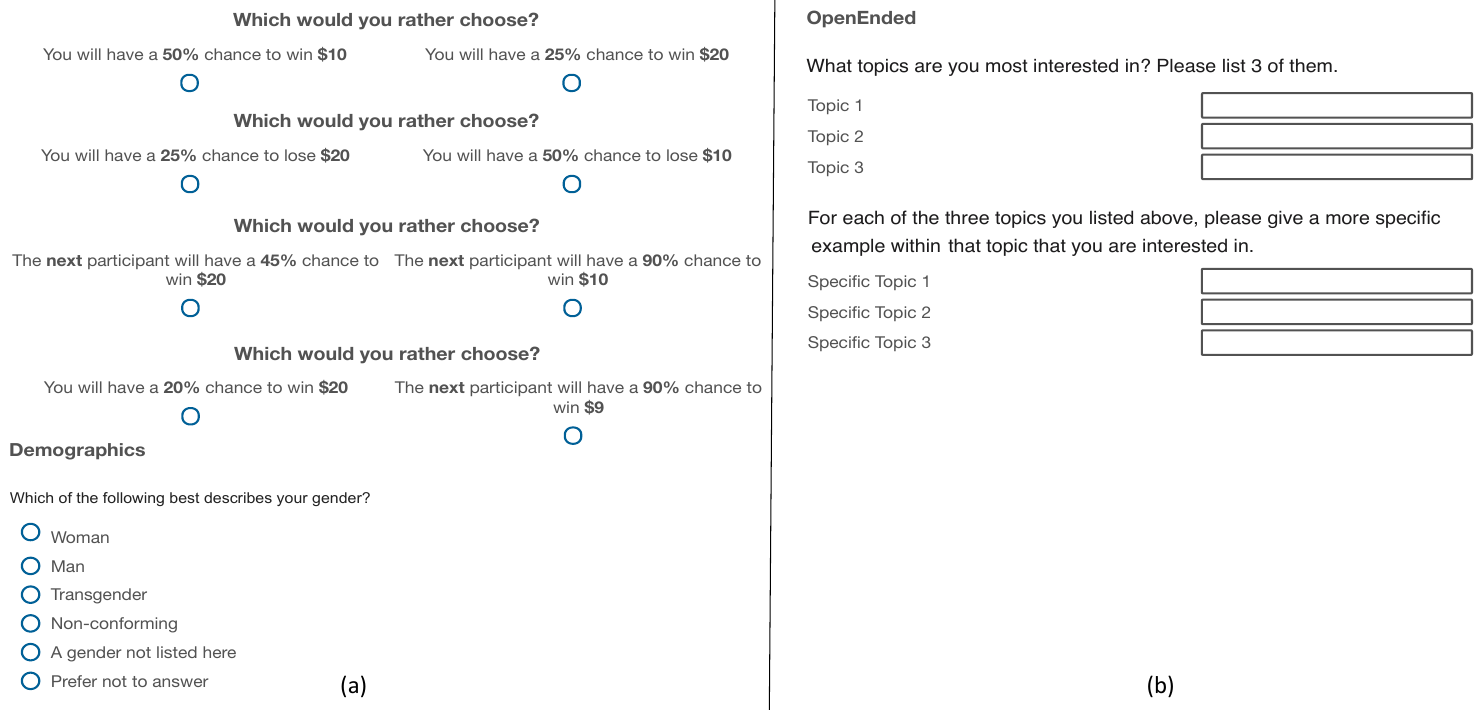}
    \caption{(a) Example of choice questions and demographic attributes and (b) the open-ended questions participant's answered with free text in the behavioral experiment.}
    \label{fig:questions}
\end{figure*}

\section{Related Work}
\label{related_work}
Representation learning has been successfully used to identify latent contexts in a high dimensional space. Pioneering research applied representation learning to words \cite{mikolov2013} and documents \cite{le2014} by converting these text-based data into vectors. In addition to natural language, representation learning is commonly used to create embeddings of images \cite{krizhevsky2012} to represent visual content and networks \cite{perozzi2014} to encode network structures.

Representation learning is also used to represent users by creating embeddings from a variety of personal data (e.g., entities relation, user attributes), and has been applied to fields such as fraud detection \cite{qli2019} and recommendation systems \cite{cheng2016} where interactions between users and a variety of entities are leveraged. Here we use representation learning to represent individual differences for choice prediction in the context of behavioral experiments. In this context, data to capture individual difference are generally collected in laboratory settings (e.g., questionnaires), where only sparse personal data are available and the sampled participants usually neither have interactions with each other nor are connected through other entities.

Unstructured user content such as user generated text in micro-blogs and online interviews have been found to be informative indicators for many personal attributes e.g., gender, age, political view \cite{rao2010} and personality \cite{jayaratne2020}. Besides predictions of personal attributes and traits, user content embedding have also been used in a variety of downstream tasks for personalization, such as text generation \cite{li2019}.
This previous research demonstrate that unstructured free-text responses can be used to capture variance among individuals. However, traditional behavioral experiments generally use free-text in qualitative analyses (e.g., anthropological fieldwork \cite{maxwell2008designing}) rather than as a quantitative tool. Converting unstructured content into a quantitative measure usually involves labor-intensive and error-prone manual coding, and is thus not commonly adopted as a measure of individual differences. In contrast to this general approach, we propose that representation learning methods can be used to quantitatively represent individual differences from unstructured user data in behavioral experiments.

\section{Dataset and Problem Definition} \label{sec:dataset}
\begin{table}[t]
\centering
\caption{The statistics of the choice dataset collected in the behavioral experiment.}
\begin{tabular}{|c|c|} 
\hline
\# of participants & 1,205                  \\
 \hline
\# of gambles & 64                  \\
 \hline
\# of gamble scenario features & 6                  \\
\hline
\# of gamble choices &77,120                  \\
 \hline
\# of words per participant           & 12                         \\ \hline
\end{tabular}
\label{tbl:dataset}
\end{table}
In this section, we first describe the behavioral experiment dataset used in this paper. The choice data is available at: \url{https://osf.io/kr8pa/}  (Experiment 1).
Each participant saw 64 choices between two gambling options. A few examples of the two gambling options are presented in Figure \ref{fig:questions} (a), e.g., ``You will have a 50\% chance to win \$10'' vs. `` You will have a 25\% chance to win \$20.'' The options differed in outcomes (e.g. \$5 vs \$10), probabilities (e.g. 50\% chance to get the outcome vs a certain outcome) and who the recipient of outcome is (self vs another participant). After participants answered the gambling questions, they answered various demographic questions and the open-ended questions analyzed here, e.g., ``what topics are you most interested in?'' to collect different aspects of personal information. The responses to the demographic questions are forced choices while the responses to the open-ended questions are free text. Figure \ref{fig:questions} (a) provides a part of the questionnaire used for this behavioral experiment. There were 1,205 participants, each responded 64 gambles, resulting 77,120 gamble choices. More details about the dataset is reported in Table \ref{tbl:dataset}. 

The outcomes, probabilities and recipients of a gambling question as well as a participant's information, ie. user identifier, degmographics or text responses to open-ended questions, are used to predict the participant's choice for the gamble question. In the following sections, we first present the approaches of learning individual embedding using user identifier (Section \ref{sec:beh2vec}) and using free text responses (Section \ref{sec:text}) for prediction. We then present a theory-based prediction model used in behavioral science for comparison (Section \ref{sec:behmodel}).

\section{Representation Learning} \label{sec:beh2vec}

\begin{figure*}[t]
    \centering
    \includegraphics[width=14cm]{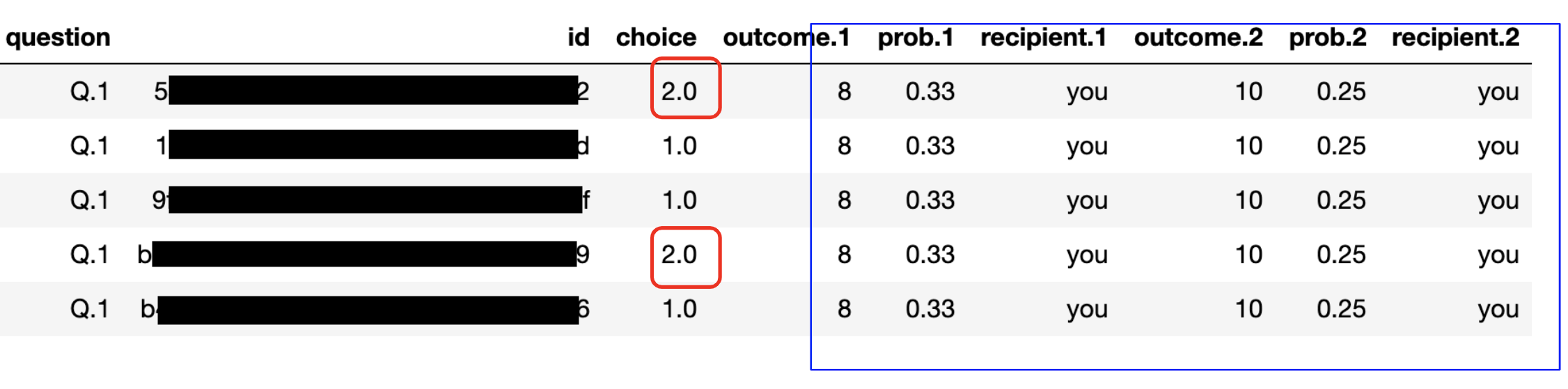}
    \caption{A user's choice can be different from another user's choice for the same scenario (gambling) features because of individual differences as appeared in the choice dataset - two of five participants chose 1 while the others chose 2. Including individual differences can reduce the confusion for prediction.}
    \label{fig:inconsis_label}
\end{figure*}

Most behavioral experiments in psychology and decision science involve a group of users answering a set of questions designed to assess their behavior in contexts of interest (e.g., preference for foods, altruistic behavior in new social contexts). The nature of these questions often depend on idiosyncratic participant preferences (e.g., food choices). As such, questions in these experiments often do not have an objectively ''correct'' response and can vary considerably between individuals (Figure \ref{fig:inconsis_label}). Although this may be reasonable from a human view point, it can be confusing from a modeling perspective as the model sees identical inputs but different outputs. It is therefore important for modeling methods to find ways to represent individual differences if they are to learn something meaningful. Here we overcome this limitations by developing a representation learning model we call \textit{Beh2vec} that represents a user by learning their decision behavior (i.e. choices made in the past). 

\begin{figure*}[htp]
    \centering
    \includegraphics[width=14cm]{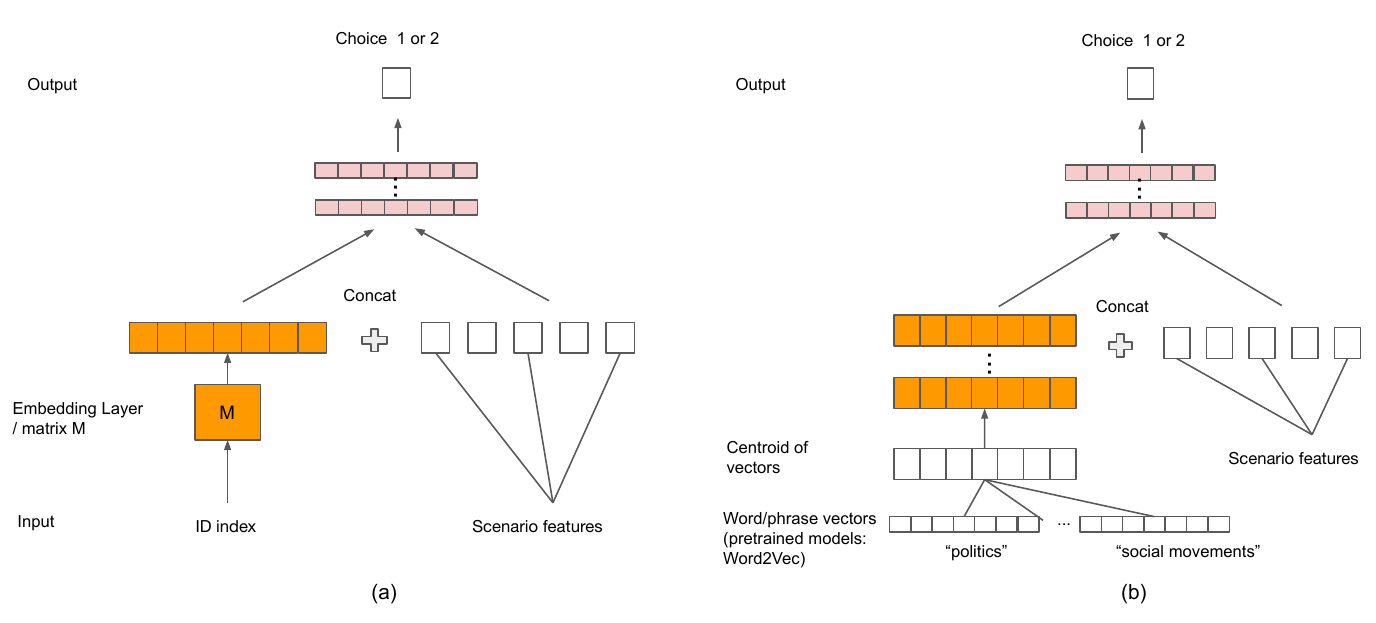}
    \caption{Model architectures - (a) the architecture of Beh2vec: an embedding layer is added to learn user-specific representations before being concatenated with the scenario features and then fed to the fully connected layers. (b) the architecture for taking inputs of free text responses uses a pre-trained embedding model Word2vec to obtain word/phrase vectors and use the centroid of these vectors as the input layer.}
    \label{fig:beh2vec}
\end{figure*}

Beh2vec, as illustrated in Figure \ref{fig:beh2vec}, is largely inspired by Word2vec \cite{mikolov2013} and Doc2vec \cite{le2014} used in Natural Language Processing. It uses other decisions made by the subject to learn the hidden embedding vectors to represent user decision behavior. We formulated the task as a binary classification problem.
More specifically, we add an embedding layer to the user identifier (“ID”) then concatenate it with the scenario features of a decision task (e.g., the 6 gamble scenario features ``outcome.1,'' ``prob.1,'' ``recepient.1,'' ``outcome.2,'' ``prob.2,'' ``recepient.2'' in Figure \ref{fig:inconsis_label} used to design gambles in the behavioral experiment in Figure \ref{fig:questions}), and feed the concatenated vector to a fully connected Multilayer Perceptron neural network for predicting a choice. 
The identifier ``ID'' token can be thought of as a contextual indicator. It acts as a memory that remembers what choices are made by a user in other decision tasks. In this framework, every user ID is mapped to a unique fixed length vector, represented by a column in a matrix $M$. The column is indexed by the index of the user in the user pool. The ID vector is shared across all contexts for the same user but not across users.

More formally, given a sequence of training decision tasks made by a user $q_1, q_2, q_3, ..., q_T$ , the objective of the user behavior vector model is to maximize the average log probability

\begin{equation} \label{eq_costfunc}
\frac{1}{T} \sum_{k=0}^{T} \log{p(C=c_k | q_1, q_2, q_3,..., q'_k,...q_T)}
\end{equation}
where $q'$ indicates that the decision task does not include the choice and we have 
\begin{equation} 
p(C=c_k | q_1, q_2, q_3,..., q'_k,... q_T) = \frac{e^{y_c}}{\sum{e^{y_{c_i}}}} 
\end{equation}
each of $y_{c_i}$ is un-normalized log-probability for each output choice $c_i$, computed as

\begin{equation} \label{eq1}
y_{c_i} = b + Uh(q_{1}, ...,q'_k,...  q_{T} ; M)
\end{equation}

where $U$, $b$ are the softmax parameters. $h$ is constructed by a concatenation or average of user behavior vectors extracted from $M$.

At training time, the user identifier embedding is trained using stochastic gradient descent via backpropagation. One can randomly sample a pair of user ID and decision tasks and compute the error gradient to update the parameters. 

At the inference time, 
one can then add more columns to $M$ to represent unseen new users and update the parameters in embedding layer with the rest parameters frozen by stochastic gradient descent using the synthetic generated data. For known users, one can identify the user embedding by looking up the embedding matrix by index. 

\section{Text Responses for Open-Ended Questions} \label{sec:text}
Representation learning as introduced in Section \ref{sec:beh2vec} can capture latent context where choices of a history of decision tasks are available. A strong advantage of representation learning is that embeddings can be created for each user based on a variety of different data sources. Demographic attributes can be used to create representation IDs, but offer only a limited representation as these are limited to a small range of categories (e.g., gender, education). Here we use open-ended questions with free-form text responses as a new measure of individual differences.

Two examples of open-ended questions in the behavioral experiment we analyzed  are presented in Figure \ref{fig:questions} (b). The open-ended questions are used to elicit user responses that provide more context for our measure of individual differences. Open-ended responses are given as free-text generated by the user. Although we did not limit the number of words users could input, we noticed that most responses were either short phrases (e.g., ``video games'' ``global warming'' ``social movements'' ) or single words (``politics'' ``sports'', ``culture''). Compared with the more restricted nature of typical demographic attributes, these free-text responses provide more diversity while still sharing some overlap between users. For example, although 13\% of users responded ``politics'' to the open-ended question ``What topics are you most interested in.'', many also followed with non-overlapping responses ( e.g., \{``politics,'' ``history,'' ``sports''\}, \{``politics,'' ``climate change,'' ``science''\}, \{``politics,'' ``finance,'' ``relationships''\}).

User responses to these open-ended questions are used to create a user-specific representation. Each word or phrase in the user's responses is converted into a vector using the pre-trained word embedding model Word2vec \cite{mikolov2013}. If any of the words in a user's responses (e.g., the 6 topics in Figure \ref{fig:questions}(b) do not appear in the Word2vec vocabulary, these words will be excluded and the rest of words in the responses will be used to learn the representation for the user. The out-of-vocabulary rate is around 10\% for the dataset used in our experiments. 

To aggregate word/phrase vectors into one vector for each user, we use the centroid of all word and phrase vectors in the user's responses to both open-ended questions as the user's vector. As presented in Figure \ref{fig:beh2vec} (b) A user vector based on the pre-trained Word2vec is then fed to a set of fully connected layers to fine-tune the representation before being concatenated with the scenario features of the decision task. Finally, it is connected to a Multilayer Perceptron that outputs choice probabilities, and the objective function in the training phase is to maximize the probabilities over all training decision tasks. 




\section{Behavioral Model} \label{sec:behmodel}
We compared our machine learning with a theory-based behavioral science subjective value  model  often used to measure individual differences on the types of gambles in the current study \cite{kahneman2013prospect,nilsson2011hierarchical}. In this task, gambles can have outcomes $V$ that result in gains or losses with probability $p$. Additionally, the result of the user's decision could either affect them (self) or another participant (other). On each trial, users chose between two gambles. The model assumes that the decision to choose either gamble is derived by computing a subjective value $SV$ of each option $i$:

\begin{equation}
    SV_{i} = p_{i} V_{i}^{\alpha_j}
\end{equation}

where $\alpha$ is a free parameter for each user $j$ that governs their aversion to risk. When $0\leq\alpha<1$, users are "risk averse" or less likely to choose uncertain gambles than prescribed by a normative strategy that maximizes return (i.e., $\alpha=1$). Conversely, when $\alpha>1$, people are measured as "risk seeking", preferring uncertain gambles more than if they were using a normative choice strategy.

People's tolerance for risk depends on choice framing, with people acting riskier when gambles are framed as losses (e.g., 50\% chance of losing \$10) than framed as gains (e.g., 50\% chance of winning \$10; \cite{kahneman2013prospect}). We modeled this by adding separate user-specific $\alpha$ parameters for gambles with losses ($\alpha_{loss}$) or gains ($\alpha_{gain}$).

To model user choices, the subjective values for each option is converted to a probability of choosing option 1 $p(c = 1)$ over option 2 using a softmax:

\begin{equation}
    p(c=1)=logit^{-1}(\beta_1 SV_{1} - \beta_2 SV_2)
\end{equation}

where $\beta_1$ and $\beta_2$ correspond to subjective value weights that depend on the recipient of the gamble (i.e., self or other). Weights given to 'self' gambles ($\beta_{self}$) and 'other' gambles ($\beta_{other}$) were treated as user-specific free parameters.

We fit this model as a multilevel hierarchical Bayesian model implemented in Stan \cite{carpenter2017stan}, which offers more precision in estimating individual level parameters than traditional fitting methods\cite{nilsson2011hierarchical}. A user $j$'s four free parameters ($\alpha_{gain}$, $\alpha_{loss}$, $\beta_{self}$, $\beta_{other}$) were estimated as a joint multivariate distribution $p(\mathbf{h_j}, \boldsymbol{\uptheta})$, with $\mathbf{h_j}$ corresponding to user-specific parameter distributions (i.e., $\mathbf{h_j} \in \{\alpha_{gains,j}$, $\alpha_{losses,j}$, $\beta_{self,j}$, $\beta_{other,j}\}$). These mutlivariate parameter distributions were assumed to be generated from a group-level prior $p(\boldsymbol{\uptheta})$. Using this prior, we sought the posterior multivariate parameter distribution for each participant, given their data $\mathbf{D_j}$:
\begin{equation}
    p(\mathbf{h_j},\boldsymbol{\uptheta}|\mathbf{D_j}) \propto p(\mathbf{D_j}|\mathbf{h_j},\boldsymbol{\uptheta})p(\boldsymbol{\uptheta})
\end{equation}

The recipient weighting parameters $\beta$ could take on any real value and were thus assumed to be generated by a Gaussian prior. The loss aversion parameters $\alpha$ can theoretically take on any positive real values. However, in practice, values of $\alpha > 1.5$ are very unlikely and allowing these larger values can result in large subjective value calculations that make fitting problematic. As such, $\alpha$ parameters were generated from a Gaussian prior that was first logit transformed to generate a number between 0 and 1 \cite{nilsson2011hierarchical} then multiplied by 1.5 to set an upper bound on the range of values that could be considered. 

We sampled from the posterior distribution using four simultaneous Hamiltonian MCMC chains of 7000 iterations with a 3500 iteration warm-up period on each chain. We fit a non-centered version of our model with a sampling step size of 0.90 to eliminate sampling divergences \cite{betancourt2015hamiltonian}. Each chain had $.99 < \hat{R} < 1.01$ and effective sample sizes $>$1/100th the number of total samples, indicating successfully chain convergence \cite{gelman1992single}.

\section{Experiments}
We first describe the experimental setting, then discuss the performance of the baselines and the models with representation learning using user IDs (Sec. \ref{sec:beh2vec}), demographic attributes and user text responses (Sec. \ref{sec:text}). We also compare our results to predictions to those from a well-validated behavioral model (Sec. \ref{sec:behmodel}) to verify whether representation learning provides a competitive measure of individual differences compared to a theory-based behavioral model. All evaluations were conducted on the behavioral experiment data introduced in Sec. \ref{sec:dataset}.


\subsection{Experimental Setting}
In the experiments, a prediction is made for a choice given the associated 6-dimensional gamble features (``outcome.1,'' ``prob.1,'' ``recepient.1,'' ``outcome.2,'' ``prob.2,'' ``recepient.2'' as shown in Figure \ref{fig:inconsis_label}) along with one of three types of personal information: user ID, demographics and text responses on two open-ended questions (Figure \ref{fig:questions}). We split the data of 77,120 gamble choices such that 80\% was used for training and 20\% used for testing. For neural network models, a two-fold cross validation was applied to the test split, i.e., one half as the validation for selecting the best model and the other half for testing, and the testing results are averaged over the folds.

Architectures of representation learning models ( Beh2vec for different personal context) used in the experiments \ref{tbl:results_repr} are described below. 
\textbf{G+UserID}: The representation learning model (Figure \ref{fig:beh2vec} (a)) uses an embedding layer of 1,205 (the number of users) $\times$ 200 (embedding size) and a dense layer for learning an ID vector. The user embedding is concatenated with the 6-dimensional gamble scenario features, then connected to a Multilayer Perceptron (MLP) with the dimension of layers: [206, 128, 64, 32, 4, 1] to predict a choice. 
\textbf{G+Text Resp.}: The representation learning model (Figure \ref{fig:beh2vec} (b)) is similar to (a) but uses pre-trained word2vec text embedding to represent words in a user's text responses. The centroid of word vectors (dimension: 300) are then fed to the network and fine-tuned with a set of fully connected layers (dimensions of layers: [300, 128, 64, 64]). The output text embedding is concatenated with the gamble scenario features, then connected to a MLP to predict a choice. The MLP has the same dimension of layers as that used in \textbf{G+UserID} except the input layer, which is replaced with a 70-dimensional layer to take a different input (64-dimensional text embedding and 6-dimensional gamble scenario features).
\textbf{G+Demographics}: The representation model similar to Figure \ref{fig:beh2vec} (b) without text embedding is used for learning representation of demographic attributes (dimension: 11) including age, gender, race, education, marital status, child, employment, income, political view, religious view and religious behavior. 

Because the choice data in the experiment are binary, we use cross-entropy loss in our implementation based on PyTorch. We set the learning rate to 0.001 for training all of the neural network models.

\subsection{Prediction Accuracy}
\begin{table*}[t] 
\centering
\caption{Testing accuracy for models using the Gambling Scenario Features (G) and three types of personal information as individual differences: User ID, Demographics and Text Responses. We report results from two machine learning approaches - Random Forest (RF) and Multilayer Perceptron (MLP), and the neural networks Beh2vec presented in Fig. \ref{fig:beh2vec} for learning representation of personal information. Among all combinations of personal context and models, the model that included text-based representations and using Beh2vec showed the highest predictive accuracy.}
\begin{tabular}{|l|l|l|l|l|}
\hline
Accuracy  & Gamble Features (G) & G+User ID                 & G+Demographics & G+Text Resp.                   \\ \hline
RF                     & 0.700           & 0.618 & 0.715 & 0.674 \\ \hline
MLP                    & 0.701           & 0.698 & 0.700 & 0.715                         \\ \hline
Beh2vec & N/A                   & 0.734    & 0.735  & \textbf{0.757}                         \\ \hline
\end{tabular}
\label{tbl:results_repr}
\end{table*}

\begin{table}[t] 
\centering
\caption{Accuracy over different types of gambling tasks where outcomes resulted in gains or losses. Accuracy from the best performing Beh2vec model (G+Text Responses) is competitive with accuracy from the theory-driven behavioral model (cf. Sec. \ref{sec:behmodel}) and even outperforms in the tasks with losses which are more challenging for the behavioral model.}
\begin{tabular}{|l|l|l|}
\hline
Accuracy  & Beh2vec w. G+Text & Behavioral Model                    \\ \hline
Gains       & \textbf{0.774}             & 0.770 \\ \hline
Losses                     & \textbf{0.748}           & 0.726 \\ \hline
All                     & \textbf{0.757}           & 0.749 \\ \hline
\end{tabular}
\label{tbl:results_gainloss}
\end{table}

We first compared the performance of MLP and Random Forest (RF) as baseline machine learning model for choice predictions. As is evident from Table \ref{tbl:results_repr}, RF is equivalent in accuracy compared to MLP when gambling features (G) are the only input. However, MLP outperforms RF when personal information is added as input. The drop in RF performance is likely due to overfiting caused by the high variability in the personal information (as indicated by RF's ~99\% training accuracy but considerably lower testing accuracy).

When the representation learning model Beh2vec (Figure \ref{fig:beh2vec}) is used to capture personal information, test accuracy improves by 3.6 \% with user ID, 3.5\% with user demographics, 4.2\% with user text responses in comparison to MLP. The improvements over different user representation models are aligned, which demonstrates the advantage of adding the representation learning layers to the neural machine learning model.

Beh2vec is most useful for text responses, which supports its capability of effectively representing less structured data. Among all combinations of personal information and models, models that included text-based representations  showed the highest predictive accuracy. Thus, although user identifiers and demographics are more commonly used in current personalized systems (e.g., recommendation), these may not capture individual differences as effectively as unstructured text-responses, particularly in behavioral experimental settings.

\subsection{Representation Learning and Behavioral Model}

Table \ref{tbl:results_gainloss} shows more details of predictions separated by whether it has a positive outcome (gain) and negative outcome (loss), which is a perspective important for behavioral experiment analysis. The results of the representation learning model Beh2vec and behavioral model shows consistency where gains predictions perform better than losses predictions. Their overall results are also competitive but provide different views of leveraging personal information in the context of choice prediction.


\subsection{Opportunities for Responsible Personalization}

We note that the prediction accuracy of the behavioral model, which only requires user IDs, approaches the accuracy of our text-based representation learning model and it is also more interpretable. Although impressive, this model has been developed over decades specifically for these types of risky gambles. However, it is limited to this one class of decisions, cannot readily account for additional task features, nor can it be made to use additional personal information (e.g., text). In contrast, our representation learning model allows including additional features and personal context that may be helpful for decision predictions in different scenarios and may need to consider the trade-off among multiple factors in responsible personalization, e.g., risk to privacy, availability, accuracy. By offering more options for metrics of individual differences, We hope that the increase of flexibility can be an aide for facilitating responsible personalization in more domains of use.
 
\subsection{Data Synthesis for Cold Start and Future Work}
Human behavior is inherently noisy \cite{Daniel2021noise} and can require large samples to enable off-the-shelf machine learning methods to produce good predictions. To address the data scarcity at cold start, one can leverage a variant of Generative Adversarial Network (GAN), Conditional Tabular GAN (CTGAN) \cite{xu2019} to synthesize human choice data. CTGAN models a tabular data distribution and samples rows from the distribution. In CTGAN, a mode-specific normalization technique is used to deal with columns that contain non-Gaussian and multimodal distributions, while conditional generator and training-by-sampling methods are used to mitigate the class imbalance problems. Table \ref{tbl:syn_data} presents preliminary experimental results of choice prediction by mixing different proportions of real and CTGAN-synthesized choice data to train the prediction model. Note that only a portion of the choice data is used in this experiment and the data split is different from the other experiments above. In the future, we will continue investigating behavioral data synthesis and other approaches such as transfer learning with learnt individual representation to tackle the challenge of data sparsity.

\begin{table}[t]
\centering
\caption{Test accuracy when training data is a mix of synthesized and real data.}
\begin{tabular}{|l|l|l|l|} 
\hline
 Training Data Size & 10,000 & 50,000                     \\ \hline
$50\% Syn.+50\% Real$   &  0.645        &  0.687                    \\ \hline
$25\% Syn.+75\% Real$   &  0.645        &  0.690                    \\ \hline
$100\% Real$ (Oracle)   & 0.672         &  0.693                    \\ \hline
\end{tabular}
\label{tbl:syn_data}
\end{table}





\section{Conclusion}
In conclusion, this paper investigates whether representation learning can help represent individual differences for choice predictions in behavioral experiments. We demonstrates the improvement of prediction accuracy over other machine learning models that do not include representation learning, and is competitive with a well-validated behavioral model. In addition, representation learning can adapt to unstructured personal information, e.g., user responses to open-ended questions, and that also results in the best performing model in our experiments. In the future, we would explore more measures of individual differences (e.g., different media, different open-ended questions or stimulus) enabled by representation learning to augment behavioral experiment analysis. 

\bibliographystyle{elsarticle-num-names}
\bibliography{sample-ceur}

\end{document}